\documentclass[conference]{IEEEtran}
\IEEEoverridecommandlockouts
\usepackage{cite}
\usepackage{amsmath,amssymb,amsfonts}
\usepackage{algorithmic}
\usepackage{graphicx}
\usepackage{textcomp}
\usepackage{multirow}
\usepackage{xcolor}
\usepackage{pgfplots}
\usepackage{booktabs}
\usepackage{subcaption}
\usepackage{pgfplots}
\pgfplotsset{compat=1.17}
\usepackage{soul}
\usepackage{array}
\usepackage{setspace}
\usepackage{float}
\usepackage{dblfloatfix}
\usepackage{subcaption}
\usepackage{array}
\usepackage{makecell}
\usepackage{multirow}
\usepackage{tikz}
\usepackage{pgfplots}
\usepackage{pgfplotstable}
\usepgfplotslibrary{statistics}
\usepackage{subcaption}
\usepackage{soul}
\usepackage{color, colortbl}
\usepackage{url}
\usepackage[normalem]{ulem}

\DeclareUnicodeCharacter{2212}{\ensuremath{-}}
\def\BibTeX{{\rm B\kern-.05em{\sc i\kern-.025em b}\kern-.08em
    T\kern-.1667em\lower.7ex\hbox{E}\kern-.125emX}}

\newcommand{\mydiamond}{%
  \sbox0{$\lozenge$}%
  \usebox0\kern-.5\wd0\clap{\raisebox{.1ex}{\scalebox{.7}[1]{$-$}}}\kern.5\wd0%
}

\begin{document}

\title{NeuraLUT-Assemble: Hardware-aware Assembling of Sub-Neural Networks for Efficient LUT Inference}

\author{\IEEEauthorblockN{Marta Andronic and George A. Constantinides}
\IEEEauthorblockA{Department of Electrical and Electronic Engineering, 
Imperial College London, UK\\
Email: \{marta.andronic18, g.constantinides\}@imperial.ac.uk}
}

\maketitle

\begin{abstract}
Efficient neural networks (NNs) leveraging lookup tables (LUTs) have demonstrated significant potential for emerging AI applications, particularly when deployed on field-programmable gate arrays (FPGAs) for edge computing. These architectures promise ultra-low latency and reduced resource utilization, broadening neural network adoption in fields such as particle physics. However, existing LUT-based designs suffer from accuracy degradation due to the large fan-in required by neurons being limited by the exponential scaling of LUT resources with input width. In practice, in prior work this tension has resulted in the reliance on extremely sparse models.

We present NeuraLUT-Assemble, a novel framework that addresses these limitations by combining mixed-precision techniques with the assembly of larger neurons from smaller units, thereby increasing connectivity while keeping the number of inputs of any given LUT manageable. Additionally, we introduce skip-connections across entire LUT structures to improve gradient flow. NeuraLUT-Assemble closes the accuracy gap between LUT-based methods and (fully-connected) MLP-based models, achieving competitive accuracy on tasks such as network intrusion detection, digit classification, and jet classification, demonstrating up to $8.42\times$ reduction in the area-delay product compared to the state-of-the-art at the time of the publication.
\end{abstract}
  
\maketitle

\section{Introduction}
Ultra-low latency NN inference has become instrumental in advancing fields such as particle physics, network security, and autonomous vehicles. In particle physics, machine learning (ML) models are essential for handling the immense data volumes generated by detectors. For instance, in the latest upgrade to the CMS trigger system at CERN's Large Hadron Collider, the system processes data from six consecutive beam crossings, every $25$ ns. The system must achieve nanosecond-level latency to be capable of accepting new data inputs continuously. Such real-time capabilities enable the collection of meaningful events that could otherwise be lost. In network security, ML-driven intrusion detection systems swiftly identify threats and anomalies, safeguarding critical infrastructure by delivering near-instant insights. Autonomous vehicles also rely on ML to interpret sensor data and make split-second navigation and safety decisions, particularly in complex or high-speed scenarios. However, due to the resource-constrained environments and strict latency requirements in these fields, deployed deep neural networks (DNNs) often fall short of the state-of-the-art accuracy in ML.

FPGAs, with their highly customizable architecture, serve as a core platform for these applications, enabling optimized computation to meet stringent performance KPIs. Their reconfigurability supports rapid design iteration, making them ideal for applications that demand frequent model updates. Moreover, FPGAs perform many computations in parallel, significantly reducing processing time.

Recent research in the field has focused on hardware-software co-design. Beyond designing efficient hardware, it is also important to adapt software for deployment efficiency. A growing area of research explores model architectures that map efficiently to hardware. For example, LUT-based approaches like NeuraLUT~\cite{NeuraLUT}, PolyLUT~\cite{poly}, LogicNets~\cite{logicnets}, NullaNet~\cite{nullanet}, PolyLUT-Add~\cite{polyadd}, AmigoLUT~\cite{amigolut}. Other recent works have adopted decision tree-based approaches like TreeLUT~\cite{treelut}, or weightless neural networks~\cite{dwn}.

Following~\cite{poly}, in this paper we refer to lookup tables of arbitrary size as Logical-LUTs (L-LUTs), highlighting their ability to exceed the capacity of the Physical-LUTs (P-LUTs) on the FPGA. When an L-LUT requires more inputs than a P-LUT can handle, logic synthesis tools map it as a circuit of multiple interconnected P-LUTs.

NeuraLUT, PolyLUT, LogicNets, and NullaNet encapsulate the entire computation of a neuron within a single L-LUT, creating a network of L-LUTs with no exposed datapaths. These prior works offer distinct trade-offs in model complexity and expressiveness. LogicNets and NullaNet excel in functional simplicity by using continuous piecewise linear functions. PolyLUT achieves similar accuracy with fewer piecewise regions (resulting in fewer L-LUTs) by employing continuous piecewise polynomial functions, delivering enhanced expressiveness but at the cost of increased training complexity. NeuraLUT trains a composition of multiple continuous piecewise linear functions, striking a balance between expressive power and reduced function complexity, leading to richer model representations for the same L-LUT complexity.

However, for any of these approaches to be feasible, the lookup table size is constrained, which can limit the accuracy of these neural networks. PolyLUT-Add~\cite{polyadd} takes a first step at trying to improve the connectivity of these networks by summing the results of multiple L-LUTs across the network. However, this approach utilizes LUTs for the implementation of the sum which are also restricted by their fan-in. AmigoLUT~\cite{amigolut} creates ensembles of smaller LUT-based NNs, including NeuraLUT, to tackle the scalability.

In this work, NeuraLUT-Assemble, we take a distinct approach to combat the challenge of the inherent limitation on the number of L-LUT inputs. We introduce a fully-parametrizable framework that assembles multiple NeuraLUT neurons as tree structures with larger fan-in, directly addressing the exponential scaling challenge. This customizable tool allows users to increase connectivity without fan-in restrictions by providing full control over the tree structure. The grouping of connections at the input of the tree structure is guided by the hardware-aware pruning strategy first introduced in the extended arXiv PolyLUT paper~\cite{polyext}.

To enhance training stability, we propose a resource-efficient method that integrates skip-connections within the L-LUTs to ensure effective gradient flow in the individual L-LUTs or across the assembled tree structure.

In summary, the novel contributions are as follows:
\begin{itemize}
    \item{We introduce NeuraLUT-Assemble, an open-source\footnote{https://github.com/MartaAndronic/NeuraLUT} toolflow that leverages the FPGA architecture by embedding dense, full-precision sub-networks within tree-structures of synthesizable Boolean lookup tables.}
    \item{We develop a fully-parametrizable framework to increase connectivity by training larger fan-in tree structures of smaller L-LUT units, where connection grouping is determined post-initial training.}
    \item{We develop a resource-efficient approach that embeds skip-connections within L-LUTs, promoting smooth gradient flow at training throughout the entire assembled tree structure.}
    \item{We assess NeuraLUT-Assemble on three standard tasks used in the low-latency DNN research community: digit classification, jet substructure classification, and network intrusion detection. Our results show that compared to NeuraLUT, our method achieves the lowest area-delay product with up to $62\times$ reduction on MNIST combined with a higher test accuracy, and up to $26\times$ reduction on jet substructure for the same test accuracy.}
\end{itemize}


\section{Background}
\label{sec:background}
Achieving real-time neural networks with low resource utilization and high accuracy requires reimagining model architecture to optimize performance on the target hardware. Recent approaches have emphasized hardware-software co-design, creating a tightly integrated cycle of model design, training, and deployment on custom hardware platforms. Co-design efforts for FPGA optimization, in particular, can be categorized by their primary learnable unit: DSP-based, XNOR-based, decision tree-based, differentiable LUT-based, and traditional LUT-based neural network architectures.

\subsection{DSP-based architectures}
In the area of ultra-low latency, \texttt{hls4ml}~\cite{duarte} is a notable open-source framework created to enable machine learning inference on FPGAs, with a focus on low-latency applications. Duarte \textit{et al.}~\cite{duarte} employ \texttt{hls4ml} to generate both fully-unrolled and rolled network architectures that target latency reduction. These designs, however, utilize high-precision networks, which results in considerable Digital Signal Processing (DSP) usage. Fahim \textit{et al.}~\cite{fahim} further optimize \texttt{hls4ml} by incorporating techniques such as quantization-aware pruning.

\subsection{XNOR-based architectures}
Ngadiuba \textit{et al.}~\cite{hls4ml} use the \texttt{hls4ml} framework to map binary and ternary neural networks onto FPGAs. Similarly, FINN~\cite{finn} is an open-source framework that was initially tailored for deploying efficient binary neural networks (BNNs) on FPGAs. To improve hardware performance, FINN replaces traditional operations with hardware-friendly alternatives, such as popcount operators instead of additions, thresholding in place of the batch normalization and activation functions, and OR gates for max-pooling.

\subsection{Decision tree-based architectures}
Decision tree-based approaches, like TreeLUT~\cite{treelut} and POLYBiNN~\cite{polybinn}, are practical methods for efficient machine learning inference on hardware platforms like FPGAs. These methods use the structure of decision trees to break down problems into smaller, manageable decisions, which work well in hardware because of their parallel and low-latency nature. TreeLUT optimizes traditional decision trees by using LUTs to speed up decision-making and reduce resource use. POLYBiNN, on the other hand, combines decision trees with polynomial regression at the leaf nodes. Both methods focus on keeping models interpretable while delivering fast inference. These approaches are useful for applications that need quick and clear decision-making, showing how classical machine learning can be adapted for modern hardware constraints.

\subsection{Differentiable LUTs}
Differentiable lookup tables offer a method to integrate neural networks with FPGA hardware by enabling gradient-based optimization of LUT configurations.
\subsubsection{LUTNet}  LUTNet~\cite{lutnet1,lutnet2}, introduced by Wang \textit{et al.}, replaces BNN XNOR operations with learned K-input Boolean functions mapped directly onto FPGA LUTs. This approach leverages the inherent flexibility of LUTs to implement complex Boolean functions, enhancing logic density and allowing for significant network pruning without accuracy degradation.

\subsubsection{Differentiable Weightless Neural Networks} A different approach, Differentiable Weightless Neural Networks (DWNs)~\cite{dwn}, utilizes an extended finite difference method to approximate gradients for training networks composed of interconnected LUTs. This technique facilitates the optimization of LUT architectures through gradient descent, despite the discrete nature of LUT outputs. In DWNs, the neural networks are constructed entirely from weightless nodes implemented as LUTs. Additionally, in DWNs a distributive thermometer encoding scheme is employed for input representation to convert continuous input features into binary vectors. However, this thermometer encoding assigns distinct floating-point thresholds to each feature, leading to potentially large overhead in converting into thermometer-encoding.

\subsection{LUT-based traditional NNs} What distinguishes this category is that, while designed for LUT-based netlist inference, the training process relies on traditional neural network models, which are later fully absorbed into LUT functions by complete enumeration. NullaNet \cite{nullanet} and LogicNets \cite{logicnets} were among the first to map entire neurons onto multi-input, multi-output Boolean functions. NullaNet minimizes these functions' footprint using Boolean logic minimization strategies and selectively determining output values for specific input combinations while treating the rest as don't-care conditions to conserve resources. In contrast, LogicNets trains neural networks that were \textit{a priori} designed to be extremely sparse to overcome NullaNet's potential accuracy loss after their don't-care optimization step. As a result of the exponential growth of the L-LUT size, LogicNets restricts the input size of each neuron to a fixed fan-in $F$, thereby keeping the truth table sizes under control to enable efficient neuron implementations. As a result, LogicNets achieves parameter complexity of $\mathcal{O}(F)$ per neuron.

PolyLUT\cite{poly} takes LogicNets' methodology further by encoding the entire neuron's function within an L-LUT but uniquely expands each neuron's feature vector to include all monomials up to a user-defined degree $D$. This gives PolyLUT a parameter complexity of $\mathcal{O}\left(\left(F+D \atop D \right)\right)$ per neuron, allowing it to capture complex data relationships through higher-degree polynomial expressions. By embedding multiplicative interactions directly within LUTs, PolyLUT avoids additional multiplication hardware, computing a continuous piecewise polynomial function—unlike LogicNets, which computes a piecewise linear function. This added complexity within each layer allows PolyLUT to reach target accuracies with fewer layers, enhancing efficiency.

Although linear functions are straightforward, PolyLUT has shown that they do not fully leverage the capacity of LUTs, making them less efficient compared to more complex functions like polynomials. However, using multivariate polynomial functions introduces exponentially growing degrees of freedom with respect to the degree of the polynomial, which complicates training and, as PolyLUT observed, offers diminishing returns when polynomial degrees exceed two~\cite{poly}.

NeuraLUT~\cite{NeuraLUT} explores an alternative universal function approximator that maintains training simplicity without requiring modifications to existing training frameworks: the multilayer perceptron (MLP)~\cite{cybenko, hornik}. By embedding MLPs within LUTs, NeuraLUT maximizes the neural network density within each individual L-LUT. This approach allows the model to achieve high expressiveness without enlarging the input size of LUTs by concentrating dense neural regions within L-LUTs and maintaining a sparse structure between them at the circuit level. However, this approach can unveil highly complex interactions between a limited number of features that is controlled by the L-LUT fan-in.

PolyLUT-Add~\cite{polyadd} represents an initial effort to enhance network connectivity by aggregating the outputs of multiple L-LUTs across the network. While this approach effectively improves feature abstraction, it relies on LUTs to perform the summation. This creates a trade-off, as the fan-in limitations of these LUTs can constrain scalability. Moreover, using LUTs for addition potentially underutilizes their capacity, as they could otherwise be leveraged for more complex operations, thereby optimizing overall network efficiency.

AmigoLUT~\cite{amigolut} ensembles multiple small models of different LUT-based NNs, such as NeuraLUT, and computes the average of the outputs of all members. This method has proven to be effective at increasing the accuracy of very weak models.

\cite{yukio} and \cite{reducedlut} focus on post-training hardware optimization, whereas our contributions are in the training process itself. Thus, these approaches are complementary, as their optimizations could be applied to our networks when the L-LUT size exceeds the P-LUT size.

\begin{figure}[tb]
\centerline{\includegraphics[width=1\columnwidth]{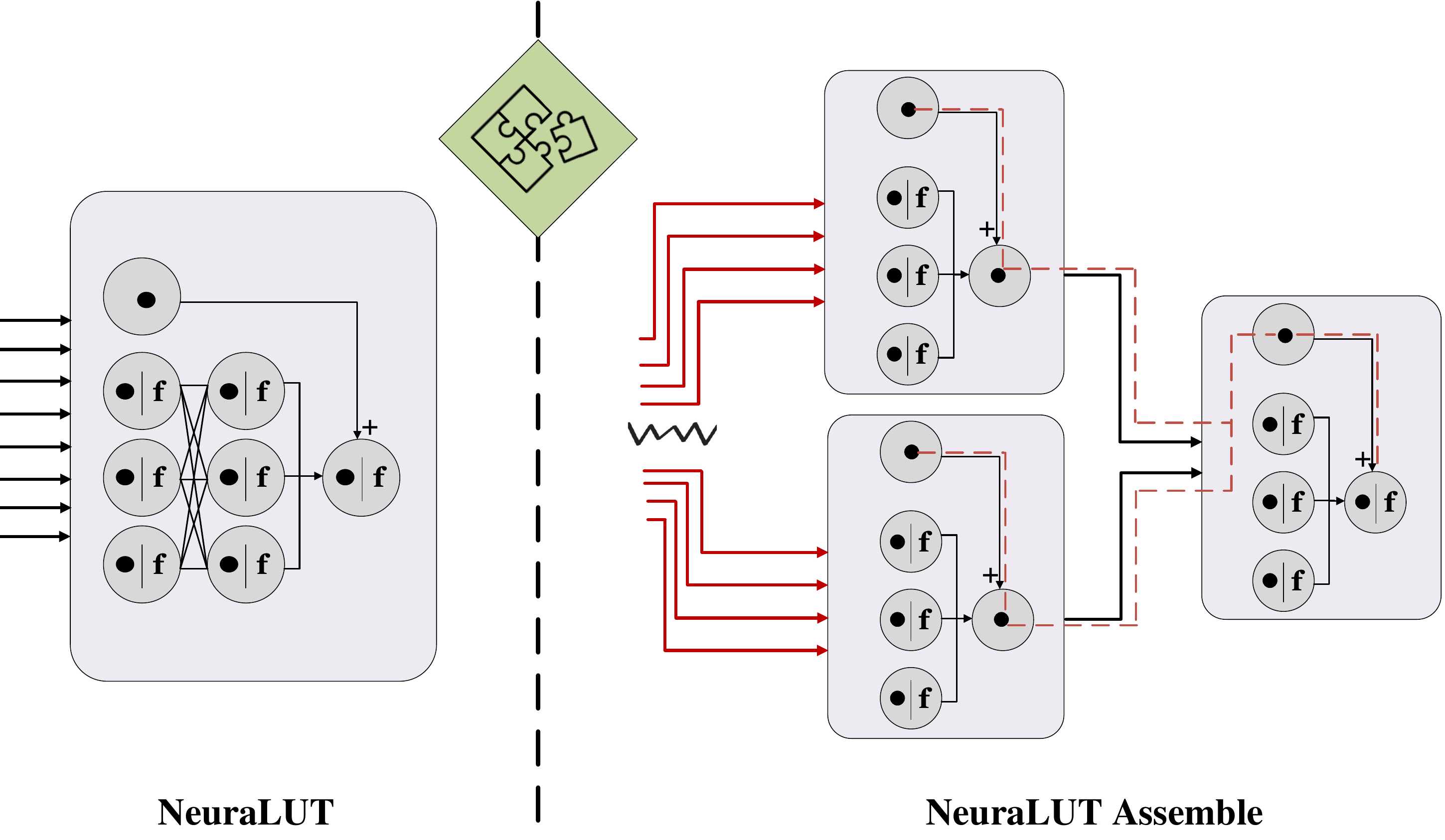}}
\caption{View of a NeuraLUT L-LUT on the left and a NeuraLUT-Assemble L-LUT tree on the right. $\bullet$ represents an affine transformation, whereas $f$ is the activation function.}
\label{fig:str}
\end{figure}

\begin{figure*}[tb]
\centerline{\includegraphics[width=1.55\columnwidth]{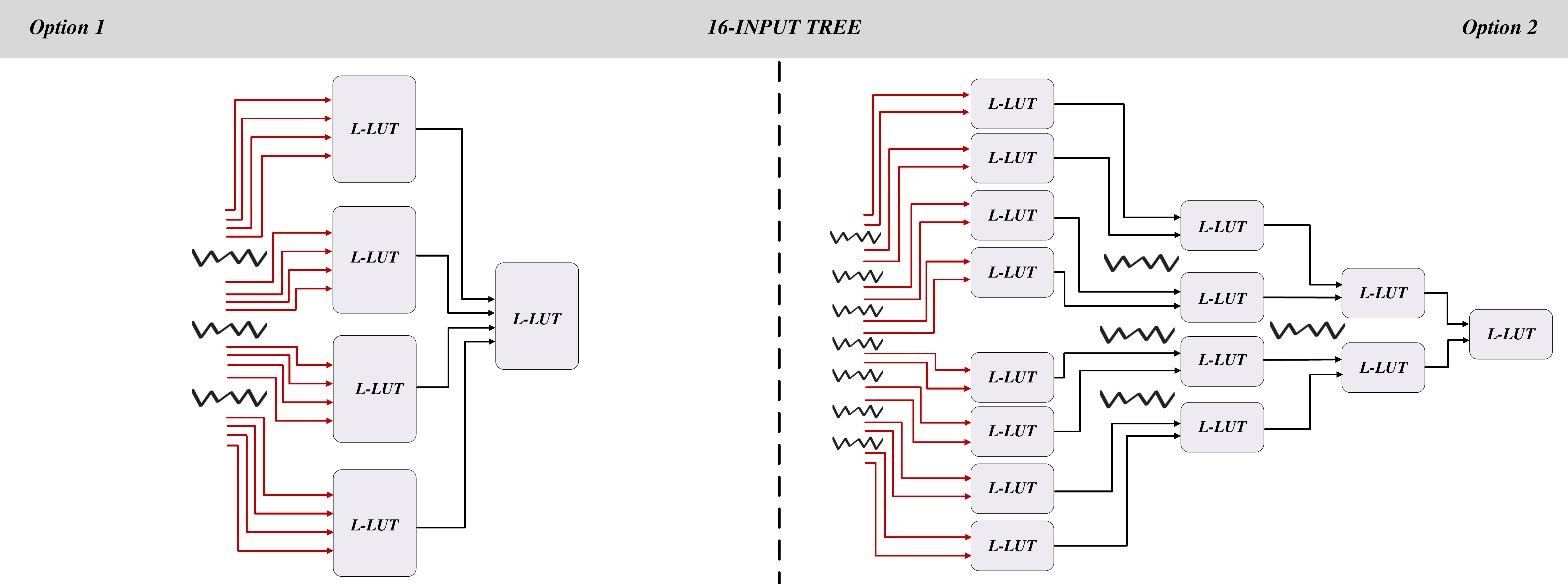}}
\caption{Example of two different NeuraLUT-Assemble configurations for a $16$-input tree. The red connections are decided upon an initial stage of dense training, whereas the black connections always stay fixed.}
\label{fig:conf}
\end{figure*}

\begin{figure}[t]
\centerline{\includegraphics[width=0.93\columnwidth]{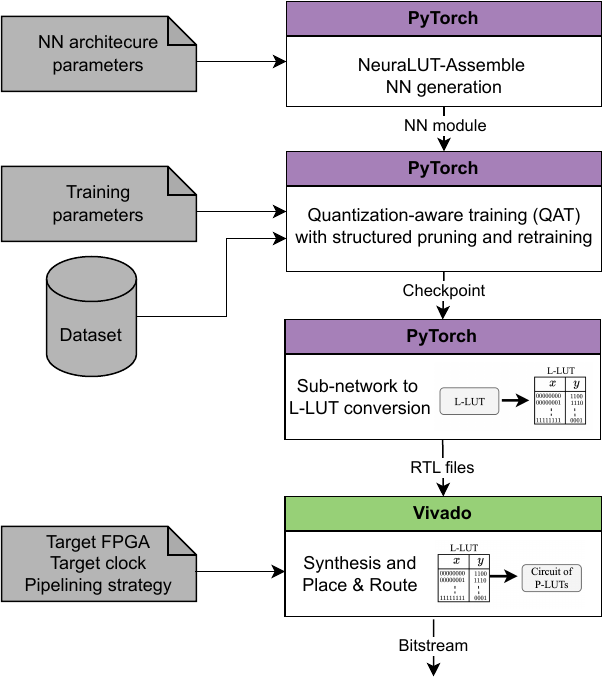}}
\caption{Visualization of NeuraLUT-Assemble's toolflow.}
\label{fig:toolflow}
\end{figure}

\subsection{Hardware-aware structured pruning}
In LUT-based neural networks, a primary challenge lies in managing the exponential increase in LUT size with a growing number of inputs. Conventional approaches address this by imposing fixed random sparsity patterns \textit{a priori}\cite{poly, NeuraLUT, logicnets}, but these methods are sensitive to initial seed selection, often resulting in performance inconsistencies.

PolyLUT~\cite{polyext} introduced a hardware-aware structured pruning strategy to overcome these limitations, promoting a tailored sparsity pattern. Rather than relying on predefined sparsity, PolyLUT defines a custom group regularizer designed to guide neuron connections according to hardware constraints.

The proposed method follows a sequential process: initially, the network undergoes dense training with a custom hardware-aware regularizer, establishing a foundation conducive to effective pruning. Following this dense training, a structured pruning stage is applied, and the resulting sparse network is subsequently retrained to restore any potential accuracy loss.

\section{Methodology} 
\label{sec:method}

Our work introduces a novel approach to designing LUT-based neural networks by leveraging a hardware-aware design to overcome the fan-in limitations of traditional LUT-based neural networks. We have designed a way to assemble tree structures of multiple L-LUTs and translate this structure onto the training framework to achieve higher connectivity. Figure~\ref{fig:str} provides a small-scale example that illustrates our strategy. Instead of training a model with $8$-input L-LUTs, we can train a fixed tree structure by combining $4$-input L-LUTs with a $2$-input L-LUT, thereby dramatically reducing the L-LUT cost. Since these two designs train fundamentally different functions, we train the tree structure from scratch rather than mapping a higher fan-in model onto it.

Training deep NNs often encounters difficulties due to the vanishing gradient problem~\cite{bengio}~\cite{glorot}, where gradients can shrink substantially as they backpropagate through layers. While this issue is less prominent in prior ultra-low latency NNs~\cite{logicnets,poly,NeuraLUT}, which tend to have limited depth, it becomes more relevant in the NeuraLUT-Assemble framework.  In NeuraLUT-Assemble, the tree-structures add additional depth to the neural network and it tends to be relatively deep compared to the number of inputs per tree, making training more difficult. To address this, we employ residual connections, which add outputs from certain layers to the activations from earlier layers, helping to preserve gradient flow~\cite{resnet}. 

A key advantage of our approach is that these residual connections traverse the entire assembled tree structure and are entirely hidden within the L-LUT synthesizable Boolean function. As a result, they do not cause an additional implementation cost or reduce regularity at inference. In NeuraLUT these connections were constrained to the L-LUT borders. As illustrated on the left side of Figure~\ref{fig:str}, a neuron without an activation function bypasses two fully connected layers, adding its output just before the activation function in the third layer. On the right side, in the NeuraLUT-Assemble strategy, all activation functions are removed except in the final tree layer, allowing a skip path with no activation function from the tree structure's input to its output, highlighted in dotted red.

\begin{figure*}[tb]
\centerline{\includegraphics[width=1.39\columnwidth]{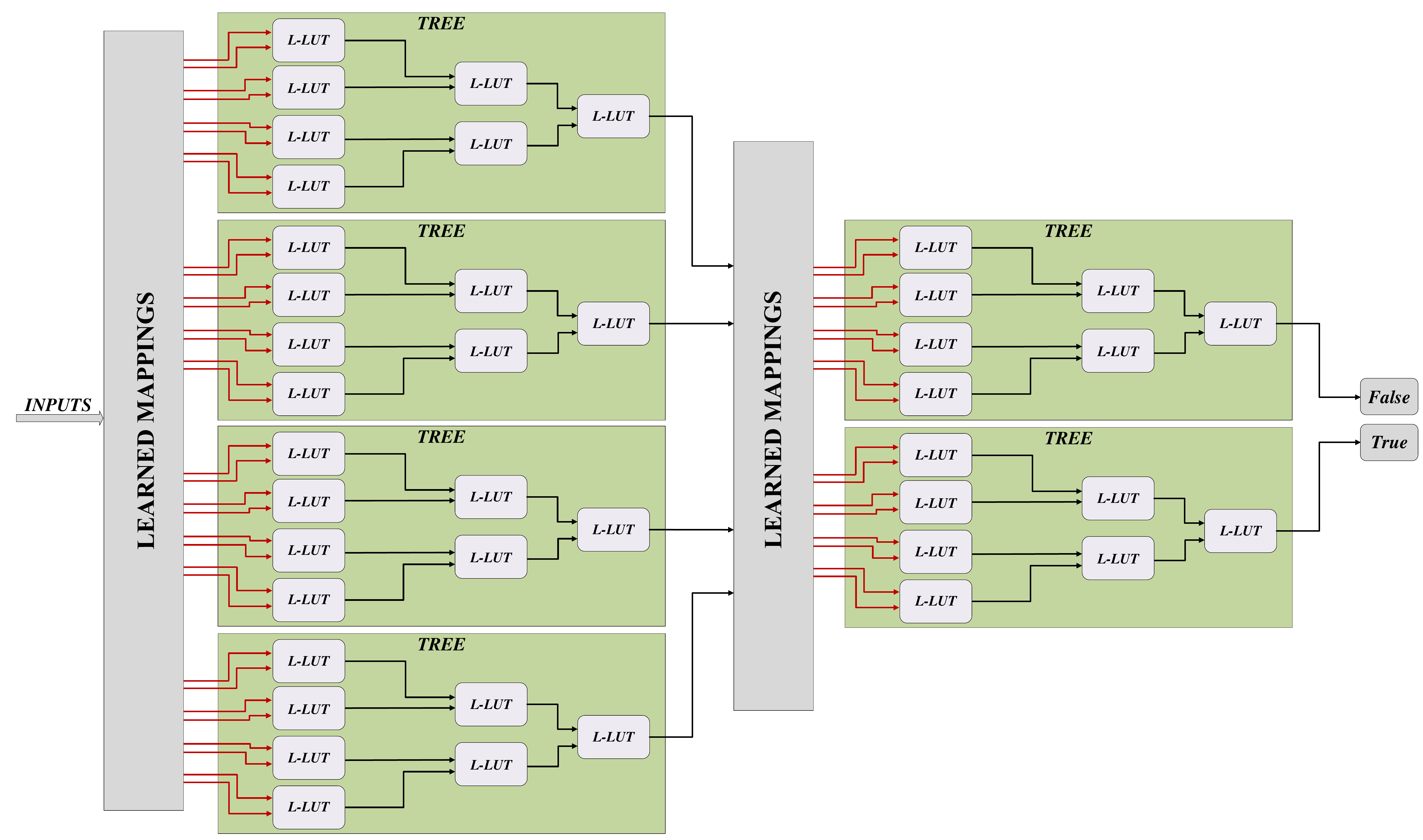}}
\caption{High-level view of toy $6$-layer NeuraLUT-Assemble network.}
\label{fig:high}
\end{figure*}

\begin{table*}[tbp]
\caption{Overview of user-defined NN architecture parameters, emphasizing the flexibility of our toolflow. The last three parameters are specific to the NN inside the L-LUT structure.}
\begin{center}
\renewcommand{\arraystretch}{1.17}
\setlength{\tabcolsep}{14pt}
\small
\begin{tabular}{rcl}
\Xhline{2\arrayrulewidth}
\textbf{Parameter} & \textbf{Notation} & \textbf{Description} \\
\Xhline{2\arrayrulewidth}
Layer sizes & $w_{l}$ & Number of L-LUTs units per layer. \\
\hline
Assemble layer & $a_{l}$ & Boolean array to indicate assemble layers, which have fixed sparsity. \\
\hline
Fan-ins & $F$ & The fan-ins are individually defined for the input, output, and inner-tree layers. \\
\hline
Bit-widths & $\beta$ & The bit-widths are individually defined for the input, output, and inner-tree layers. \\
\hline
\rowcolor[gray]{0.9} Depth of L-LUT NN & $L$ & Depth parameter for the NN hidden inside the L-LUTs. \\
\hline
\rowcolor[gray]{0.9} Width of L-LUT NN & $N$ & Width parameter for the NN hidden inside the L-LUTs. \\
\hline
\rowcolor[gray]{0.9} Skip-connection step & $S$ & Skip-connection step for the NN hidden inside the L-LUTs. \\
\end{tabular}
\renewcommand{\arraystretch}{1.2}
\label{table:parameters}
\end{center}
\end{table*}

\subsection{Assembling strategies}
Our fully customizable framework allows users to construct tree structures tailored to their needs, balancing the trade-off between the number of L-LUTs and their size. Figure~\ref{fig:conf} illustrates two possible configurations for a tree with $16$ inputs. In the first configuration, each L-LUT has $4$ inputs, and a number of entries in the LUT of $2^{4\beta}$, where $\beta$ is the number of activation bits. In the second configuration, the number of L-LUTs increases threefold, but the size of each L-LUT decreases to the square root of its previous value, \textit{i.e.} to $2^{2\beta}$, as the number of inputs per L-LUT is halved. Networks trained using the NeuraLUT-Assemble strategy resemble the structure shown in Figure~\ref{fig:high}.

This exponential reduction in size has significant implications for hardware implementation. This strategy enhances scalability, reducing the size of L-LUTs directly. However, the trade-off lies in the increased number of L-LUTs required to construct the tree, which can lead to higher interconnect complexity and potentially longer critical paths in the hardware. Therefore pipelining requires more attention here than in previous LUT-based NN work and we experiment with different strategies in Section~\ref{fig:high}.

There is also an inherent trade-off in training between the expressivity of the model and its hardware cost. Larger input L-LUTs have higher expressivity because they can capture more complex inter-dependencies among the input features. However, as the tree structure becomes deeper, the individual functions implemented by the L-LUTs at each layer become simpler. However, this simplicity makes the model easier and faster to train. To balance these competing factors, we allow the input connections to be decided upon an initial stage of training. This approach enables the framework to prioritize and preserve the most critical inter-dependencies among inputs. By doing so, the model retains a high degree of expressivity while mitigating the challenges of training and the hardware cost associated with larger L-LUTs. We investigate this tradeoff empirically in Section~\ref{sec:evaluation}.

\begin{table*}[ht]
\centering
\caption{Reference floating-point (FP), fully-connected (FC) test accuracy, our test accuracy and architecture parameters for different models used for evaluation in Table~\ref{table:evaluation_combined}. Despite the sparsity and low precision, NeuraLUT-Assemble achieves accuracy comparable to that of a dense floating-point model.}
\small
\renewcommand{\arraystretch}{1.1}
\setlength{\tabcolsep}{5pt}
\begin{tabular}{lcccccccccc}
\toprule
\textbf{Dataset} & \multicolumn{2}{c}{\textbf{Accuracy (\%)}} & \multicolumn{7}{c}{\textbf{Parameters}} \\
\cmidrule(lr){2-3} \cmidrule(lr){4-10}
 & \textbf{FP FC} & \textbf{Ours} & \textbf{$w_{l}$} & \textbf{$a_{l}$} & $F$ & \textbf{$\beta$} & \textbf{$L$} & \textbf{$N$} & \textbf{$S$} \\
\midrule
\multirow{2}{*}{\textbf{MNIST(+aug/-aug)}} & 98.7\% & 98.6\%  & [2160,360,2160,360,60,10] & [0,1,0,1,1,1] & [6,6,6,6,6,6] & [1,1,1,1,1,6] & 2 & 64 & 2 \\
                                & 98.4\% & 97.9\% & [2160,360,2160,360,60,10] & [0,1,0,1,1,1] & [6,6,6,6,6,6] & [1,1,1,1,1,6] & 2 & 64 & 2 \\
\midrule
\multirow{1}{*}{\textbf{JSC CERNBox}} & 76.0\% & 75.0\% & [320,160,80,40,20,10,5] & [0,1,1,1,1,1,1] & [1,2,2,2,2,2,2] & [8,4,4,4,4,4,8] & 2 & 64 & 2 \\
\midrule
\multirow{1}{*}{\textbf{JSC OpenML}} & 77.0\% & 76.0\% & [320,160,80,40,20,10,5] & [0,1,1,1,1,1,1] & [1,2,2,2,2,2,2] & [6,3,3,3,3,3,8] & 2 & 64 & 2 \\
\midrule
\multirow{1}{*}{\textbf{NID}} & 92.5\% & 93.0\% & [60,20,9,3,1] & [0,1,0,1,1] & [6,3,3,3,3] & [1,2,2,2,2,2] & 2 & 16 & 2 \\
\bottomrule
\end{tabular}
\label{tab:fp_accuracy_parameters}
\end{table*}
\begin{table*}[ht]
\centering
\caption{Latency, F\textsubscript{max}, LUTs, and FFs for pipeline every 3 L-LUT layers and pipelining every L-LUT layer. The results are taken after running \texttt{Out-of-Context} synthesis and place \& route, matching with prior work.}
\small
\renewcommand{\arraystretch}{1}
\begin{tabular}{lcccccccc}
\toprule
\textbf{Dataset} & \multicolumn{4}{c}{\textbf{Pipelining every L-LUT layer}} & \multicolumn{4}{c}{\textbf{Pipelining every 3 L-LUT layers}} \\
\cmidrule(lr){2-5} \cmidrule(lr){6-9}
& \textbf{Latency (ns)} & \textbf{F\textsubscript{max} (MHz)} & \textbf{LUTs} & \textbf{FFs} & \textbf{Latency (ns)} & \textbf{F\textsubscript{max} (MHz)} & \textbf{LUTs} & \textbf{FFs} \\
\midrule
\multirow{2}{*}{\textbf{MNIST(+aug/-aug)}} & 6.5 & 916 & 5040 & 5464 & 2.2 & 849 & 5037 & 713 \\
                                & 6.6 & 912 & 5089 & 5699 & 2.1 & 863 & 5070 & 725 \\
\midrule
\multirow{1}{*}{\textbf{JSC CERNBox}} & 7.0 & 994 & 8535 & 2717 & 5.7 & 352 & 8539 & 1332 \\
\midrule
\multirow{1}{*}{\textbf{JSC OpenML}} & 6.6 & 1067 & 1844 & 1983 & 2.1 & 941 & 1780 & 540 \\
\midrule
\multirow{1}{*}{\textbf{NID}} & 3.4 & 1479 & 95 & 187 & 1.4 & 1471 & 91 & 24 \\
\bottomrule
\end{tabular}
\label{tab:latency_fmax_area}
\end{table*}

\subsection{Toolflow}
NeuraLUT-Assemble builds upon the NeuraLUT toolflow~\cite{NeuraLUT}, enabling seamless DNN training, conversion into L-LUTs, RTL file generation, and hardware compilation and verification. The training implementation has been modified to accommodate the unique hidden NNs and the tree-based structure of NeuraLUT-Assemble. A high-level overview of the toolflow stages is presented in Fig.~\ref{fig:toolflow}.

\subsubsection{Quantization-aware training (QAT)}
The training code is written in PyTorch. Initially, the user needs to specify the learning parameters (\textit{e.g.}, learning rate) and the topology parameters, as detailed in Table~\ref{table:parameters}. The hyperparameters $L$, $N$, and $S$ define the structure of the sub-networks within the L-LUTs as detailed in Table~\ref{table:parameters}. The rest of the hyperparameters, such as $w_l$, $a_l$, $F$, and $\beta$ dictate the tree-level topology.

Once these parameters are defined and the dataset is selected, the model is trained using Decoupled Weight Decay Regularization~\cite{reg} and Stochastic Gradient Descent with Warm Restarts~\cite{sgd}. Each sub-network incorporates batch normalization and quantization through Brevitas~\cite{brevitas}, which applies learned scaling factors to the activation functions. The training process spans $1000$ epochs for jet tagging and $500$ epochs for MNIST and network intrusion.

\subsubsection{Sub-network to L-LUT conversion}
After training, each sub-network within the tree is transformed into an L-LUT. This conversion is automatically performed in PyTorch by generating all possible input combinations based on the specified bit-widths and evaluating the corresponding output values through inference. The number of entries in each L-LUT is $2^{\beta F}$, similar to LogicNets, but with differences in the specific lookup table content derived from the sub-network functions.

\subsubsection{RTL file generation}
Using the PyTorch framework, the trained network is automatically converted into Verilog RTL, where each L-LUT is implemented as a read-only memory (ROM) block.

\subsubsection{Synthesis and Place \& Route}
To synthesize and compile the generated Verilog RTL files, we use Vivado $2020.1$, targeting the \texttt{xcvu9p-flgb2104-2-i} FPGA. Consistent compilation settings are applied, including Vivado's \texttt{Flow\_PerfOptimized\_high} mode and \texttt{Out-of-Context} synthesis, which allows the user to synthesize a design module independently of any other parts of the design. This is the standard way to isolate the delay and size of a computational FPGA core, enabling direct comparison with \cite{logicnets,polyext,NeuraLUT,polyadd, treelut, dwn, amigolut}. In practice, the maximum clock frequencies are limited by the FPGA global clock, however, the very high frequencies achievable by NeuraLUT-Assemble demonstrate that this computational core is highly unlikely to be on the critical path in any larger, high-speed system. The target clock period is set depending on the size of the network.

\subsection{Pipelining strategies}
The user can choose to prioritize latency by adding a register after every three L-LUT layers or opt for a throughput-optimized design by placing a register after each L-LUT layer. Both strategies have been analyzed, as shown in Table~\ref{tab:latency_fmax_area}. To further enhance performance, we enabled the Vivado \texttt{retiming} option.

\section{Evaluation} 
\label{sec:evaluation}
\begin{table*}[tb]
\caption{Evaluation of NeuraLUT against other ultra-low latency neural networks, with results from cited conference papers. Our results are taken after running \texttt{Out-of-Context} synthesis and place \& route, matching with prior work.}
\centering
\renewcommand{\arraystretch}{1.18}
\setlength{\tabcolsep}{5pt}
\small
\begin{tabular}{clrrrrrrrr}
\toprule
\textbf{Dataset} & \textbf{Model} & \textbf{Accuracy} & \textbf{LUT} & \textbf{FF} & \textbf{DSP} & \textbf{BRAM} & $\mathbf{F_\text{max}}$ & \textbf{Latency} & \textbf{Area$\times$Delay} \\
& &\textbf{(\%)}& & & & & \textbf{(MHz)} & \textbf{(ns)} & \textbf{(LUT$\times$ns)} \\
\midrule
\multirow{10}{*}{\textbf{MNIST}} 
& \textbf{NeuraLUT-Assemble}$^\mathbf{+aug}$ & \textbf{98.6\%} & 5037 & 713 & \textbf{0} & \textbf{0} & 849 & 2.2 & $1.11\times10^4$\\
& \textbf{NeuraLUT-Assemble} & 97.9\% & 5070 & 725 & \textbf{0} & \textbf{0} & 863 & \textbf{2.1} & $\mathbf{1.06\times10^4}$ \\
& TreeLUT~\cite{treelut} & 96.6\% & 4478 & \textbf{597} & \textbf{0} & \textbf{0} & 791 & 2.5 & $1.12\times10^4$ \\
& DWN~\cite{dwn} & 97.8\% & \textbf{2092} & 1757 & \textbf{0} & \textbf{0} & 873 & 9.2 & $1.92\times10^4$ \\
& PolyLUT-Add~\cite{polyadd} & 96\% & 14810 & 2609 & \textbf{0} & \textbf{0} & 625 & 10 & $1.48\times10^5$ \\
& AmigoLUT-NeuraLUT~\cite{amigolut} & 95.5\% &  16081 & 13292 & \textbf{0} & \textbf{0} & \textbf{925} & 7.6 & $1.22\times10^5$ \\
& NeuraLUT~\cite{NeuraLUT} & 96\% & 54798 & 3757 & \textbf{0} & \textbf{0} & 431 & 12 & $6.58\times10^5$ \\
& PolyLUT~\cite{polyext} & 97.5\% & 75131 & 4668 & \textbf{0} & \textbf{0} & 353 & 17 & $1.38\times10^6$ \\
& FINN~\cite{finn} & 96\% & 91131 & \textemdash & \textbf{0} & 5 & 200 & 310 & $2.82\times10^7$ \\
& \texttt{hls4ml} (Ngadiuba et al.)~\cite{hls4ml} & 95\% & 260092 & 165513 & \textbf{0} & 345 & 200 & 190 & $4.94\times10^7$ \\
\midrule
\multirow{5}{*}{\textbf{JSC CERNBox}} 
& \textbf{NeuraLUT-Assemble} & 75.0\% & \textbf{8539} & 1332 & \textbf{0} & \textbf{0} & 352 & \textbf{5.7} & $\mathbf{4.87\times10^4}$ \\
& AmigoLUT-NeuraLUT~\cite{amigolut} & 74.4\% &  42742 & 4717 & \textbf{0} & \textbf{0} & \textbf{520} & 9.6 & $4.10\times10^5$ \\
& PolyLUT-Add~\cite{polyadd} & 75\% & 36484 & \textbf{1209} & \textbf{0} & \textbf{0} & 315 & 16 & $5.84\times10^5$ \\
& NeuraLUT~\cite{NeuraLUT} & 75\% & 92357 & 4885 & \textbf{0} & \textbf{0} & 368 & 14 & $1.29\times10^6$ \\
& PolyLUT~\cite{polyext} & \textbf{75.1\%} & 246071 & 12384 & \textbf{0} & \textbf{0} & 203 & 25 & $6.15\times10^6$ \\
& LogicNets~\cite{logicnets} & 72\% & 37931 & 810& \textbf{0} & \textbf{0} & 427 & 13 & $4.93\times10^5$ \\
\midrule
\multirow{4}{*}{\textbf{JSC OpenML}}
& \textbf{NeuraLUT-Assemble} & 76.0\% & \textbf{1780} & 540 & \textbf{0} & \textbf{0} & \textbf{941} & \textbf{2.1} & $\mathbf{3.92\times10^3}$ \\
& TreeLUT~\cite{treelut} & 75.6\% & 2234 & \textbf{347} & \textbf{0} & \textbf{0} & 735 & 2.7 & $6.03\times10^3$ \\
& DWN~\cite{dwn} & \textbf{76.3\%} & 6302 & 4128 & \textbf{0} & \textbf{0} & 695 &  14.4 & $9.07\times10^4$ \\
& \texttt{hls4ml} (Fahim et al.)~\cite{fahim} & 76.2\% & 63251 & 4394 & 38 & \textbf{0} & 200 & 45 & $2.85\times10^6$ \\
\midrule
\multirow{5}{*}{\textbf{NID}} 
& \textbf{NeuraLUT-Assemble} & \textbf{93.0\%} & \textbf{91} & \textbf{24} & \textbf{0} & \textbf{0} & \textbf{1471} & \textbf{1.4} & $\mathbf{1.27\times10^2}$ \\
& TreeLUT~\cite{treelut} & 92.7\% & 345 & 33 & \textbf{0} & \textbf{0} & 681 & 1.5 & $5.17\times10^2$ \\
& PolyLUT-Add~\cite{polyadd} & 92\% & 1649 & 830 & \textbf{0} & \textbf{0} & 620 & 8 & $1.32\times10^4$ \\
& PolyLUT~\cite{polyext} & 92.2\% & 3165 & 774 & \textbf{0} & \textbf{0} & 580 & 9 & $2.85\times10^4$ \\
& LogicNets~\cite{logicnets} & 91\% & 15949 & 1274 & \textbf{0} & \textbf{0} & 471 & 13 & $2.07\times10^5$ \\
\bottomrule
\end{tabular}
\label{table:evaluation_combined}
\end{table*}
\definecolor{paleRed}{RGB}{255, 204, 204}
\definecolor{paleGreen}{RGB}{153, 204, 153}
\definecolor{paleGreenDarker}{RGB}{102, 153, 102}
\begin{figure*}[t] 
    \centering
    \begin{tikzpicture}

        \begin{axis}[
            width=14cm, height=6cm,
            bar width=20pt,
            ylabel={\textcolor{paleGreenDarker}{Area (LUTs)}},
            ymin=0, ymax=9000,
            ylabel near ticks,
            xtick={1,2,3,4,5,6,7,8,9},
            xticklabel style={rotate=0, text width=11mm, font=\scriptsize, align=center},
            xtick=\empty,
            yticklabel style={color=paleGreenDarker},
            yticklabel style={/pgf/number format/.cd, fixed, fixed zerofill, precision=0},
            legend style={at={(0.5,1.1)}, anchor=south, legend columns=1},
            axis y line*=left,
            ymajorgrids,
        ]
            \addplot[ybar, paleGreenDarker, fill = paleGreenDarker, xshift=5pt] coordinates {(1, 6515)};
            \addplot[ybar, paleGreenDarker, fill = paleGreenDarker, xshift=3pt] coordinates {(2, 250)};
            \addplot[ybar, paleGreenDarker, fill = paleGreenDarker, xshift=1pt] coordinates {(3, 1929)};
            
           \addplot[ybar, paleGreenDarker, fill = paleGreenDarker] coordinates {(4, 6515) (5, 250) (6, 1929)};
            \addplot[ybar, paleGreenDarker, fill = paleGreenDarker, xshift=-2pt] coordinates {(7, 6515)};
            \addplot[ybar, paleGreenDarker, fill = paleGreenDarker, xshift=-3pt] coordinates {(8, 250)};
            \addplot[ybar, paleGreenDarker, fill = paleGreenDarker, xshift=-5pt] coordinates {(9, 1929)};
            
        \end{axis}
    
        \begin{axis}[
            width=14cm, height=6cm,
            axis y line*=right,
            ylabel={\textcolor{purple}{Test Accuracy (\%)}},
            ylabel near ticks,
            ymin=71.5, ymax=76,
            xtick={1,2,3,4,5,6,7,8,9},
            xticklabel style={rotate=0, text width=11mm, font=\scriptsize, align=center},
            xticklabels={{\textbf{(1)}\\\textbf{4-input 2-depth}}, {\textbf{(2)}\\\textbf{2-input 4-depth}}, {\textbf{(3)}\\\textbf{2-input 6-depth}}, {\textbf{(1)}\\\textbf{4-input 2-depth}}, {\textbf{(2)}\\\textbf{2-input 4-depth}}, {\textbf{(3)}\\\textbf{2-input 6-depth}}, {\textbf{(1)}\\\textbf{4-input 2-depth}}, {\textbf{(2)}\\\textbf{2-input 4-depth}}, {\textbf{(3)}\\\textbf{2-input 6-depth}}},
            yticklabel style={color=purple},
            boxplot/draw direction=y,
            boxplot/box extend=0.01cm,
            boxplot/every box/.style={solid, fill=paleRed, draw=purple, line width=0.5pt},
            boxplot/every whisker/.style={solid, purple, line width=0.5pt},
            boxplot/every median/.style={solid, purple, line width=0.5pt}
        ]
            \addplot+[
                boxplot prepared={
                    median=74.22771,
                    upper quartile=74.39,
                    lower quartile=74.16265,
                    upper whisker=74.40482,
                    lower whisker=74.14759
                }
            ] coordinates {};

            \addplot+[
                boxplot prepared={
                    median=73.89578,
                    upper quartile=73.96,
                    lower quartile=73.85663,
                    upper whisker=73.93795,
                    lower whisker=73.84458
                }
            ] coordinates {};

            \addplot+[
                boxplot prepared={
                    median=74.33976,
                    upper quartile=74.47,
                    lower quartile=74.34578,
                    upper whisker=74.48916,
                    lower whisker=74.33855
                }
            ] coordinates {};
            
            \addplot+[
                boxplot prepared={
                    median=73.9357,
                    upper quartile=74.0866,
                    lower quartile=73.6283,
                    upper whisker=74.1801,
                    lower whisker=72.5259
                }
            ] coordinates {};

            \addplot+[
                boxplot prepared={
                    median=73.01747,
                    upper quartile=73.4506,
                    lower quartile=72.84458,
                    upper whisker=73.67048,
                    lower whisker=71.72651
                }
            ] coordinates {};

            \addplot+[
                boxplot prepared={
                    median=74.23193,
                    upper quartile=74.35964,
                    lower quartile=74.21205,
                    upper whisker=74.41627,
                    lower whisker=74.0994
                }
            ] coordinates {};
            
            \addplot+[
                boxplot prepared={
                    median=74.13916,
                    upper quartile=74.3506,
                    lower quartile=74.08193,
                    upper whisker=74.38373,
                    lower whisker=74.18012
                }
            ] coordinates {};

            \addplot+[
                boxplot prepared={
                    median=73.75241,
                    upper quartile=73.88916,
                    lower quartile=73.6988,
                    upper whisker=73.88916,
                    lower whisker=73.41145
                }
            ] coordinates {};

            \addplot+[
                boxplot prepared={
                    median=74.05422,
                    upper quartile=74.08133,
                    lower quartile=73.89458,
                    upper whisker=74.16928,
                    lower whisker=73.9006
                }
            ] coordinates {};

        \node[draw, fill=white, rounded corners, align=center, font=\scriptsize] at (axis cs:2, 75.5) {\textbf{Complete Model}};
        \node[draw, fill=white, rounded corners, align=center, font=\scriptsize] at (axis cs:5, 75.5) {\textbf{w/o Learned Mappings}};
        \node[draw, fill=white, rounded corners, align=center, font=\scriptsize] at (axis cs:8, 75.5) {\textbf{w/o} \textbf{Tree-Level Skips}};

        \draw[black, dashed, thick] (axis cs:3.45,\pgfkeysvalueof{/pgfplots/ymin}) -- (axis cs:3.45,\pgfkeysvalueof{/pgfplots/ymax});
        \draw[black, dashed, thick] (axis cs:6.5,\pgfkeysvalueof{/pgfplots/ymin}) -- (axis cs:6.5,\pgfkeysvalueof{/pgfplots/ymax});
            
        \end{axis}
        
    \end{tikzpicture}
    \caption{Comparison of area (green bar plot) and test accuracy (red box plot) for three architecture options on the JSC dataset. The options are as follows: (1) a $16$-input tree using $4$-LUTs with $2$-depth (corresponding to Option 1 in Figure~\ref{fig:conf}), (2) a $16$-input tree using 2-LUTs with 4-depth (corresponding to Option 2 in Figure~\ref{fig:conf}), and (3) a $64$-input tree using $2$-LUTs with $6$-depth (obtained by extending Option 2 with two additional layers). Each architecture was evaluated under three configurations: Complete Model (NeuraLUT-Assemble), without Learned Mappings (using random connectivity), and without Tree-Level Skip-Connections (retaining only intra-L-LUT connections). These configurations introduce negligible variability in area, as the architecture size remains fixed.
    }
    \label{fig:area-accuracy}
\end{figure*}
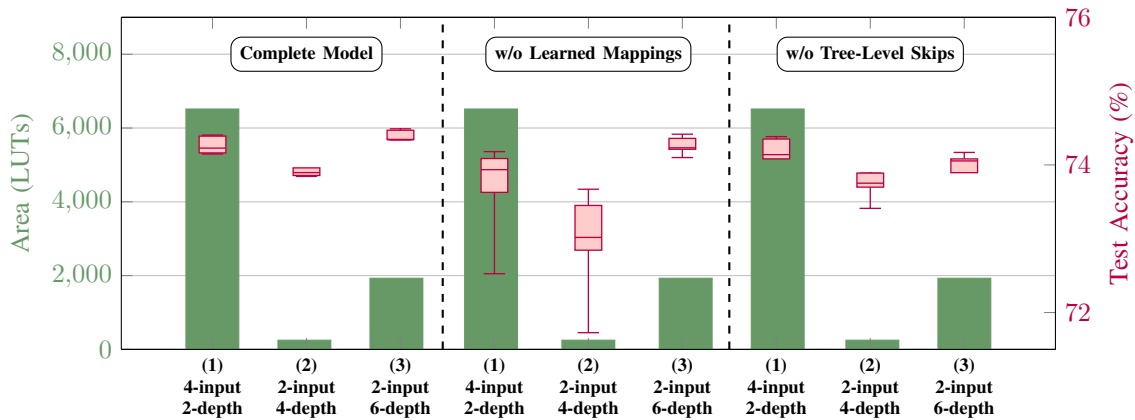
We evaluate NeuraLUT-Assemble on three common tasks used in the ultra-low latency research community. The first one is the digit classification task using the MNIST~\cite{mnist} dataset. The MNIST dataset consists of handwritten digits represented as $28\times28$ pixel images. The input pixels are flattened, resulting in inputs of size $784$, while the $10$ output classes correspond to each digit. The second one is the jet substructure classification (JSC), which can be performed using two different datasets. Both datasets target the same jet classification task, but the CERNBox version \cite{CERNboxDataset} contains $986806$ instances, while the OpenML version \cite{OpenMLHLF} has about $830000$ instances. The discrepancy likely arises from different filtering or selection criteria. Experimentally, we observed that models trained on the OpenML dataset achieve higher accuracy, suggesting that its smaller size reflects more stringent filtering or better data curation. This particular task involves processing $16$ substructure properties to classify $5$ types of jets. The last one is the network intrusion task using the UNSW-NB15 dataset as utilized in \cite{murovic}. This dataset is designed for the purpose of labelling network packets as either safe ($0$) or malicious ($1$), based on $49$ input features.

\subsection{Comparison with prior work}
To ensure a fair comparison with prior works, we selected parameters and architectures that either match or exceed the test accuracy of other ultra-low-latency approaches. We evaluated metrics such as area, latency, maximum frequency, and the area-delay product. Table~\ref{tab:fp_accuracy_parameters} summarizes the parameters used, along with a reference for the test accuracy achieved using the same-sized network with floating-point precision and fully-connected layers. Notably, none of our models deviate by more than $1$ percentage point in test accuracy compared to these references.

Table~\ref{table:evaluation_combined} provides a comprehensive comparison, showing that NeuraLUT-Assemble outperforms all other prior works in terms of the area-delay product. Compared to the best performing prior work from each dataset, we show reductions of $1.06\times$, $8.42\times$, $1.54\times$, $4.07\times$, respectively.

For MNIST, we evaluated two models: one without data augmentation to align with prior works, and another incorporating data augmentation. Compared to NeuraLUT, our method achieves a $62\times$ reduction in the area-delay product while improving test accuracy by $2$ percentage points, underscoring the efficiency of our approach. Notably, while NeuraLUT relied on $12$-input LUTs to reach this level of accuracy, NeuraLUT-Assemble achieves even higher accuracy with only $6$-input LUTs, thus the exponential area reduction. Furthermore, compared to LUT-based approaches aimed at improving connectivity, such as PolyLUT-Add~\cite{polyadd} and AmigoLUT-NeuraLUT~\cite{amigolut}, we observe $14\times$ and $11.5\times$ reductions in the area-delay product, respectively, while also achieving approximately $2$ percentage points higher test accuracy. When compared to the decision-based TreeLUT approach~\cite{treelut}, NeuraLUT-Assemble offers comparable hardware performance and delivers $2$ percentage point higher accuracy. Similarly, against weightless neural networks, NeuraLUT-Assemble achieves a $1.8\times$ improvement in the area-delay product for comparable accuracy.

For the JSC dataset from CERNBox, we focused on comparisons with other LUT-based approaches, as they are the only prior works utilizing this data source. Here, NeuraLUT-Assemble demonstrates a $26\times$ reduction in the area-delay product while maintaining the same test accuracy as NeuraLUT. On the OpenML datasource, NeuraLUT-Assemble achieves at least a $1.5\times$ reduction in the area-delay product compared to prior works.

On the network intrusion dataset (NID), compared to TreeLUT, NeuraLUT-Assemble reduces the area-delay product by $4\times$ for higher accuracy. The frequency discrepancy is most likely due to our extra pipeline stages, compared to TreeLUT which only has one. Compared to PolyLUT-Add, NeuraLUT-Assemble reduces the area-delay product by $104\times$, while also increasing test accuracy by $2$ percentage points. The NID dataset has $593$ one-bit inputs and it is likely that only a small subset of these inputs is truly relevant for classification. Methods like \cite{logicnets, poly, polyadd} randomly select neuron fan-in, potentially wasting logic on less informative bits. In contrast, NeuraLUT-Assemble learns and efficiently selects the most relevant inputs, optimizing resource usage.

\subsection{Pipelining study}
Table~\ref{tab:latency_fmax_area} presents an analysis of two pipelining strategies: adding a register after every three L-LUT layers versus after each L-LUT layer. On MNIST, JSC OpenML, and NID, where the L-LUT size matches the size of the P-LUTs, the difference in $F_{max}$ between the two configurations remains within $1.15\times$, while the $3$-layer pipelining strategy reduces the total cycles threefold. However, for the JSC CERNBox dataset, the larger L-LUTs (synthesized as circuits of P-LUTs) result in greater circuit depth, which reduces the achievable $F_{max}$ by nearly $3\times$. The users can tailor their designs based on application-specific requirements. High-frequency applications benefit from per-layer pipelining, while latency-focused designs are better served by $3$-layer pipelining.

\subsection{Jet substructure ablation study}
\label{sec:study}

In Figure~\ref{fig:area-accuracy}, we present a study on the JSC dataset showcasing the primary advantage of our methodology: assembling highly efficient NNs using neurons with strict fan-in constraints. Figure~\ref{fig:conf} illustrates the transition from a $16$-input tree constructed with $4$-input LUTs (Option 1) to a $16$-input tree constructed with $2$-input LUTs (Option 2). Our results, based on models labeled (1) and (2), show that transitioning from Option 1 to Option 2 reduces the area by $26\times$ while incurring less than a $0.5$ percentage point drop in test accuracy. This demonstrates our methodology's ability to maintain connectivity while achieving a significant reduction in area footprint. Extending the Option 2 architecture with two additional layers using the same $2$-input LUTs (model labeled (3)) recovers the accuracy loss and even surpasses the accuracy of (1), while still maintaining a reduced area footprint, now by $3.4\times$.

We also conducted an ablation study by removing the learned mapping stages, as reflected in the models labeled ``w/o Learned Mappings" in Figure~\ref{fig:area-accuracy}. This removal not only degraded the overall test accuracy of all model options but also increased result variability based on the random seed. This effect was less pronounced for (3), which has $4\times$ more input-layer connections, thereby reducing its sensitivity to suboptimal input connections.

In the models labeled ``w/o Tree-Level Skips," we ablated the tree-level skip-connections. Notably, the tree depths for (1), (2), and (3) are $2$, $4$, and $6$, respectively. The results indicate that the accuracy drop caused by removing skip-connections increases with tree depth, emphasizing their importance for robust training, particularly in deeper tree architectures.

\section{Conclusion and Future Work}
\label{sec:furtherwork}
NeuraLUT-Assemble advances LUT-based neural networks by addressing fan-in limitations and resource constraints through a flexible, hardware-aware framework. Our results demonstrate significant reductions in the area-delay product while maintaining competitive accuracy across benchmarks, outperforming prior approaches in efficiency and scalability. 

By enabling larger fan-in structures and leveraging skip-connections for robust training, NeuraLUT-Assemble provides customizable solutions for low-latency or high-throughput applications. This work highlights the potential of hardware-aware AI designs and sets the stage for further optimization and broader applicability. Future work will focus on fine-grained optimization of framework parameters and explore the creation of hybrid networks that combine the strengths of different architectures to enhance performance.

\clearpage
\bibliographystyle{IEEEtran}
\begingroup
\raggedright
\bibliography{bibs}
\endgroup

\end{document}